\documentclass[preprint,10pt]{elsarticle}

\usepackage[english]{babel}
\usepackage[T1]{fontenc}
\usepackage[a4paper,top=2.5cm,bottom=2cm,left=2.5cm,right=2.5cm]{geometry}
\usepackage{amsmath, graphicx, fancyhdr, tabularx, makecell}
\usepackage[colorinlistoftodos]{todonotes}
\usepackage[colorlinks=true, allcolors=blue]{hyperref}
\usepackage{natbib}
\usepackage{multirow}
\usepackage{svg}
\usepackage{pdflscape}

\makeatletter
\def\ps@pprintTitle{%
  \let\@oddhead\@empty
  \let\@evenhead\@empty
  \let\@oddfoot\@empty
  \let\@evenfoot\@empty
}
\makeatother 

\begin{document}

\begin{frontmatter}

\title{Toward AI-Driven Phytoplankton Event Detection from Image Classification: Temporal Aggregation Under Domain Shift}

\author[1]{Ali Rohan\corref{cor1}}
\author[2]{Hassan Shayan}
\author[2]{Han-Gue Jo}
\author[3]{Colin Moffat}

\address[1]{%
Wolfson School of Mechanical, Electrical and Manufacturing Engineering,
Loughborough University,
Loughborough LE11 3TU, UK
}
\address[2]{Department of AI Convergence, College of Computer and Software, Kunsan National University, Gunsan 54150, Republic of Korea}
\address[3]{Robert Gordon University, Garthdee House, Garthdee Rd, Aberdeen AB10 7AQ, UK}

\begin{abstract}
Imaging FlowCytobot (IFCB) systems generate large volumes of phytoplankton imagery that increasingly exceed the capacity of manual expert analysis. Although machine-learning approaches have achieved high image-level classification performance, it remains unclear whether such predictions can be translated into ecologically meaningful temporal information suitable for monitoring applications. This study investigates whether AI-derived phytoplankton classifications can reproduce temporal occurrence patterns and elevated-occurrence signals under realistic sampling conditions and independent-domain transfer. A monitoring-oriented framework was developed using MedPlanktonSet and evaluated using an independent IFCB dataset from the Swedish Meteorological and Hydrological Institute (SMHI). Eight biologically comparable taxa were selected as monitoring targets, and all remaining observations were retained as a background category to preserve the full image-stream denominator. Session-aware partitioning was used to prevent temporal leakage, and alternative preprocessing strategies and convolutional neural network architectures were evaluated. Cross-domain experiments showed that preprocessing decisions strongly influenced generalisation. Under the final nine-class framework, EfficientNet-B0 provided the strongest target-versus-background discrimination and was selected for temporal analysis. Despite residual image-level errors, AI-derived target-occurrence rates closely reproduced expert-derived reference signals after aggregation (mean absolute error = 0.0432, Pearson $r$ = 0.9696, Spearman $\rho$ = 0.9568). Strong agreement was maintained across multi-session windows, and elevated, strong, and persistent occurrence proxies were recovered in held-out testing. These findings demonstrate that AI-based phytoplankton classification can support monitoring-relevant temporal indicators beyond image-level identification. Temporal aggregation transformed predictions into robust occurrence signals, supporting AI-assisted phytoplankton monitoring while highlighting the need for temporally continuous datasets and independent event annotations for future operational early-warning systems.
\end{abstract}

\begin{keyword}
Phytoplankton Monitoring \sep Imaging FlowCytobot (IFCB) \sep Deep Learning \sep Temporal Ecological Indicators \sep Ecological Early-Warning Systems 
\end{keyword}

\end{frontmatter}

\section{Introduction}
Phytoplankton are fundamental components of marine ecosystems, sustaining pelagic food webs, regulating biogeochemical cycling, and contributing substantially to global carbon export \citep{1,2,3}. Their ecological importance also creates a significant monitoring challenge, as some taxa can increase rapidly and generate harmful or disruptive events that affect aquaculture, fisheries, tourism, public health, and coastal management \citep{4,5,6}. Consequently, monitoring programmes require more than taxonomic inventories; they require timely information on whether ecologically relevant taxa are increasing, persisting, or becoming dominant at a rate that may warrant further investigation or intervention \citep{7,8,9}.

Traditional microscopy remains the taxonomic reference standard because it combines species-level identification with expert ecological interpretation. However, manual analysis is labour-intensive and difficult to scale to the temporal resolution needed for rapidly changing plankton communities \citep{10,11}. Automated imaging technologies such as FlowCAM, ZooScan, underwater imaging systems, and the Imaging FlowCytobot (IFCB) have therefore become increasingly important within marine observing programmes \citep{10,12,13,14}. Among these, IFCB is particularly relevant for phytoplankton monitoring because it continuously images nano- and microplankton in flow, enabling high-frequency observations of community dynamics and potentially harmful taxa \citep{7,8,13}. While such systems alleviate the sampling bottleneck, they simultaneously create a new challenge: transforming large image streams into ecologically meaningful information \citep{11,15}.

Advances in machine learning have substantially improved automated plankton classification. Early approaches relied on hand-crafted descriptors of morphology, texture, and shape combined with conventional classifiers \citep{16,17,18}. More recently, Convolutional Neural Networks (CNNs) have become the dominant paradigm by learning discriminative visual features directly from labelled images \citep{19,20}. Numerous studies have demonstrated strong image-level performance for plankton recognition and have contributed valuable datasets, deep-learning architectures, and operational processing workflows. Progress in the field has included the development of large-scale benchmark datasets for fine-grained plankton classification \citep{21}, the successful application of convolutional neural networks to automated plankton image analysis \citep{22}, the incorporation of contextual metadata to improve recognition performance \citep{23}, the design of efficient deep-learning systems for in situ deployment \citep{24}, and the integration of automated image-classification workflows into phytoplankton monitoring applications \citep{25}. Collectively, these advances have substantially improved the feasibility of large-scale automated plankton analysis. Nevertheless, evaluation remains focused primarily on image-level metrics such as accuracy, precision, recall, F1-score, and confusion matrices. While these measures provide important assessments of recognition performance, they offer limited insight into whether classifier outputs remain meaningful after temporal aggregation or whether they preserve the ecological patterns required for monitoring applications.

Image-level performance, however, does not directly translate into monitoring utility. A classifier assigns labels to individual images; it does not directly identify ecological events or population dynamics. Ecological interpretation emerges only after image-level predictions are aggregated across a sampling unit and examined through time. Consequently, errors introduced at the image level may be amplified, dampened, or systematically biased during aggregation, potentially influencing the temporal patterns inferred from the resulting data stream \citep{26,27}. A classifier that performs well on conventional benchmark metrics may therefore not necessarily preserve the temporal characteristics required for ecological interpretation. This distinction is particularly important for operational monitoring systems, where the ultimate objective is not image recognition itself, but the generation of temporally meaningful information about changes in community composition.

Previous studies provide evidence that image-derived plankton time series can support ecological analysis. Automated imaging has been used to characterise community variability, monitor elevated harmful taxa, and investigate responses to changing environmental conditions \citep{7,8,9}. Furthermore, Faillettaz et al. \citep{26} demonstrated that useful ecological patterns can emerge even when image-level classification is imperfect. These findings suggest that monitoring value may depend less on absolute classification accuracy and more on whether aggregated predictions preserve ecologically relevant temporal behaviour. Despite this, relatively few studies explicitly evaluate classifier outputs as temporal ecological signals rather than as independent image-level predictions.
A second challenge concerns model transferability. Plankton image datasets differ in instrumentation, optical characteristics, segmentation procedures, object orientation, object size distributions, community composition, and annotation practices \citep{11,28}. Such differences give rise to domain shift, a well-recognised problem in machine learning whereby models trained on one data distribution often experience performance degradation when applied to another \citep{29,30,31}. Recent plankton studies have confirmed that domain shift represents a significant obstacle to operational deployment and have highlighted the need for robust cross-dataset evaluation strategies \citep{32,33}. Yet the relationship between domain shift and the preservation of temporal ecological signals remains insufficiently explored.

The unresolved problem, therefore, lies at the intersection of three active research areas. First, automated phytoplankton classification has advanced substantially, yet image-level performance alone does not establish monitoring utility. Second, image-derived time series can support ecological interpretation, but this requires explicit temporal aggregation and evaluation of the resulting signal. Third, domain shift is increasingly recognised as a barrier to deployment, yet its consequences are typically examined independently of temporal monitoring objectives. Consequently, it remains unclear whether AI-derived phytoplankton classifications can preserve temporally meaningful ecological patterns after aggregation while maintaining acceptable performance when applied outside the training domain.

This study addresses these questions using labelled IFCB image collections from MedPlanktonSet \citep{34} and an independent IFCB dataset provided by the Swedish Meteorological and Hydrological Institute (SMHI)\citep{35}. Rather than proposing a new CNN architecture, the study evaluates a complete pipeline that transforms image-level classifications into temporally aggregated ecological indicators. Classifier predictions are aggregated into session-level target-occurrence rates, compared with expert-label-derived temporal patterns, and analysed across fixed-length multi-session windows to characterise persistence and temporal change. In parallel, external evaluation is performed on an independent regional dataset to assess cross-domain transferability.

The contributions of this work are fourfold. First, it evaluates phytoplankton classification from the perspective of ecological signal generation rather than image recognition alone. Second, it employs session-aware training, validation, and test partitioning to minimise temporal leakage and provide a realistic assessment of monitoring performance. Third, it compares temporal signals constructed from expert-labelled images with those constructed from AI predictions to determine whether image-level classifications preserve ecologically interpretable patterns after temporal aggregation. Fourth, it quantifies the effects of domain shift through external evaluation on an independent IFCB dataset. Because independent bloom ground-truth annotations are unavailable, the study focuses on temporal indicators, target-occurrence rates, event proxies, and bloom-like dynamics rather than claiming direct bloom detection. Collectively, these analyses provide a framework for evaluating whether AI-based phytoplankton classification can support ecological monitoring under realistic conditions of temporal sparsity and cross-regional variability.
 
The remainder of this paper is organised as follows. Section 2 describes the datasets, target-taxon selection process, temporal framework, image-classification methodology, and the procedures used for temporal-indicator and event-proxy construction. Section 3 presents the experimental results, including analytical dataset characteristics, input-representation and model-development analyses, image-classification performance, temporal signal reconstruction, event-proxy validation, and cross-domain evaluation. Section 4 discusses the implications of the findings, including challenges related to temporal monitoring, event validation, and cross-domain generalisation, together with considerations for future operational deployment. Finally, Section 5 concludes the paper and summarises the main contributions of the proposed monitoring-oriented framework.

\section{Methodology}
\subsection{Study overview}
This study was designed to evaluate whether image-level phytoplankton classifications can be transformed into temporally meaningful ecological indicators under realistic monitoring constraints. The proposed framework integrates image classification, temporal aggregation, event-proxy analysis, and external validation using two timestamped Imaging FlowCytobot (IFCB) datasets. MedPlanktonSet \citep{34} was used for model development, in-domain evaluation, and temporal signal analysis, whereas the Swedish Meteorological and Hydrological Institute (SMHI) IFCB dataset \citep{35} was reserved exclusively for independent external evaluation.

\begin{figure}
\centering
\includegraphics[width=1\textwidth]{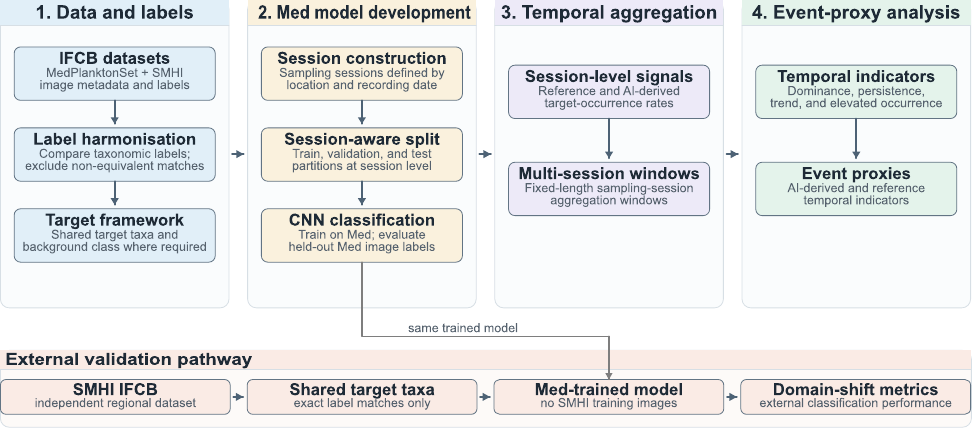}
\caption{Workflow of the AI-based phytoplankton temporal-monitoring framework.}
\label{fig:1}
\end{figure}

The overall workflow is illustrated in Figure \ref{fig:1}. First, image-level metadata from MedPlanktonSet and SMHI were organised into a unified indexing framework containing taxonomic labels, timestamps, locations, dataset identifiers, and image paths. Second, the label spaces of the two datasets were compared to identify biologically defensible shared target taxa. Non-equivalent taxonomic labels were excluded to ensure that cross-dataset comparisons were performed only on comparable biological classes. Third, images were mapped to a target-label framework, including a background (OTHER) class where required for the calculation of target-occurrence rates.
Subsequently, timestamped MedPlanktonSet images were organised into recording sessions, defined as all images acquired at the same location on the same recording day. Session-aware train–validation–test partitions were then constructed to prevent images from the same sampling session appearing in multiple subsets. Convolutional neural network (CNN) classifiers were trained using MedPlanktonSet and evaluated using held-out image data. Although image-level classification performance was assessed using conventional recognition metrics, the primary objective of the study was to determine whether image-level predictions could preserve ecologically meaningful temporal patterns after aggregation.

To address this objective, reference and AI-derived temporal signals were constructed by aggregating expert-labelled observations and AI-generated classifications using the same session framework. These signals were subsequently summarised across fixed-length sampling-session aggregation windows and analysed using temporal indicators describing target occurrence, dominance, persistence, and temporal change. Event proxies were then derived from the aggregated signals and compared to evaluate the degree to which AI-generated classifications reproduced patterns observed in the corresponding reference signals.

Finally, the Med-trained model was applied to the independent SMHI IFCB dataset using exact shared target taxa only. Because no SMHI images were used during model development, this evaluation provided an independent assessment of cross-domain transferability under geographically distinct sampling conditions. As independent bloom annotations were unavailable, the study focuses on temporal indicators, target-occurrence signals, event proxies, and bloom-like dynamics rather than attempting direct bloom detection.

\subsection{Dataset selection, metadata preparation, and label harmonisation}
Two Imaging FlowCytobot (IFCB) image datasets were selected to support both temporal signal analysis and independent cross-domain evaluation. MedPlanktonSet \citep{34} was selected as the primary dataset because it provides expert-labelled phytoplankton images together with timestamps and sampling-location information, enabling image-level observations to be linked to sampling events through time. The dataset comprised 77,271 labelled images collected between November 2022 and February 2025 from six locations within the Gulf of Naples. These characteristics allowed MedPlanktonSet to be used for both image-classification model development and the construction of temporally aggregated reference signals. The principal characteristics of both datasets are summarised in Table \ref{tab:1}. An independent IFCB dataset provided by the Swedish Meteorological and Hydrological Institute (SMHI) \citep{35} was used to assess the transferability of the resulting models beyond the training domain. The SMHI collection contained 82,122 labelled images acquired between August 2016 and October 2025 from three monitoring regions in Swedish and adjacent waters. To ensure a fully independent evaluation, SMHI images were not used during model training, validation, model selection, hyperparameter optimisation, session construction, or temporal-signal generation. Their role was restricted to external evaluation of model performance under different geographical, environmental, and annotation conditions.
For both datasets, image-level metadata were organised into a common indexing framework containing dataset identifiers, taxonomic labels, sampling locations, timestamps, and image paths. Records lacking the information required for the relevant stage of analysis were excluded during data preparation. Establishing a unified metadata structure ensured that identical processing, aggregation, and evaluation procedures could be applied consistently across both datasets while preserving their distinct analytical roles within the study design.

\begin{table*}[t]
\centering
\caption{Summary of the datasets used in this study.}
\label{tab:1}
\small
\begin{tabularx}{\textwidth}{l|X|X|c|c|c}
\hline
Dataset & Role in study & Region/Locations & Time span & Images & Taxonomic classes \\
\hline
MedPlanktonSet &
Model development, in-domain testing, and temporal aggregation &
Gulf of Naples (LTER-MC, Torre del Greco, Sarno River, Napoli Porto, Ammontatura, and Capri Canyon) &
2022-11-24 to 2025-02-25 &
77,271 &
139 \\ \hline

SMHI IFCB &
Independent external cross-domain evaluation only &
Baltic Sea, Skagerrak--Kattegat, and Tångesund &
2016-08-10 to 2025-10-10 &
82,122 &
145 \\
\hline
\end{tabularx}
\end{table*}

A key challenge was that the two datasets were developed independently and therefore used different taxonomic label spaces. MedPlanktonSet contained 139 original labels, whereas the SMHI dataset contained 145 labels. An initial comparison identified 19 labels that appeared to be shared between the two datasets. However, apparent similarity in label names did not necessarily imply biological equivalence because taxa were sometimes annotated at different taxonomic resolutions or using different naming conventions. Consequently, the objective was not to maximise the number of shared classes, but rather to identify a subset of taxa that could be compared reliably and consistently across datasets.

Candidate target taxa were selected using two criteria. First, the label had to represent the same taxonomic concept in both datasets without requiring subjective relabelling or merging of biologically distinct categories. Second, the taxon had to be sufficiently represented in both datasets to support meaningful model development and evaluation, with a minimum requirement of 50 images in each dataset.

Labels that did not satisfy these criteria were excluded. For example, MedPlanktonSet contained annotations such as Thalassionema-spp and Thalassiosira-gravida, whereas the SMHI dataset contained labels including Thalassiosira-spp-single, Thalassiosira-spp-chain, and Thalassiosira-punctigera. Although these labels refer to related centric diatoms, they represent different genera, species-level annotations, or morphology-specific categories and therefore could not be considered equivalent classes for direct cross-dataset evaluation. Such labels were excluded to avoid introducing uncertainty into subsequent analyses.

Application of the selection criteria reduced the initial set of 19 apparently shared labels to eight target taxa: Chaetoceros-spp, Cryptophyceae, Cerataulina-pelagica, Pyramimonas-spp, Dactyliosolen-fragilissimus, Chaetoceros-danicus, Chaetoceros-peruvianus, and Dictyocha-fibula (Table \ref{tab:2}). These taxa represented the largest biologically comparable subset that could be evaluated consistently across both datasets while maintaining sufficient image support for robust analysis.
For temporal analyses, all MedPlanktonSet observations that did not belong to the selected target taxa were assigned to a background OTHER category. This step was necessary because temporal indicators were derived from target-occurrence rates calculated relative to the complete image stream. Including the OTHER category ensured that target-occurrence rates reflected the proportion of target observations within the full phytoplankton community rather than within a preselected subset of images, thereby reducing the risk of artificially inflating target-occurrence estimates. 

\begin{table*}[t]
\centering
\caption{Target taxa selected following taxonomic harmonisation between MedPlanktonSet and the SMHI IFCB dataset.}
\label{tab:2}
\small
\begin{tabular}{l|r|r|c|c}
\hline
Target taxon &
Med images &
SMHI images &
Med framework &
Strict SMHI evaluation \\
\hline
\textit{Chaetoceros} spp. &
12,695 &
6,240 &
Yes &
No \\ \hline

Cryptophyceae &
2,022 &
3,215 &
Yes &
No \\ \hline

\textit{Cerataulina pelagica} &
956 &
390 &
Yes &
Yes \\ \hline

\textit{Pyramimonas} spp. &
1,383 &
292 &
Yes &
No \\ \hline

\textit{Dactyliosolen fragilissimus} &
220 &
1,189 &
Yes &
Yes \\ \hline

\textit{Chaetoceros danicus} &
880 &
201 &
Yes &
Yes \\ \hline

\textit{Chaetoceros peruvianus} &
540 &
144 &
Yes &
No \\ \hline

\textit{Dictyocha fibula} &
577 &
62 &
Yes &
Yes \\ 
\hline

Total &
20,273 &
11,733 &
8 taxa &
4 taxa \\
\hline

\end{tabular}
\end{table*}

The eight-taxon framework was used for model development, in-domain evaluation, and temporal analyses. For external validation, evaluation was restricted to four taxa (Cerataulina-pelagica, Dactyliosolen-fragilissimus, Chaetoceros-danicus, and Dictyocha-fibula) that possessed directly corresponding labels in both datasets (Table \ref{tab:2}). The remaining target taxa were retained within the MedPlanktonSet framework but were not included in the strict SMHI evaluation because equivalent SMHI labels were unavailable or not directly comparable. Restricting external evaluation to exact taxonomic matches ensured that cross-domain performance estimates reflected genuine model transferability rather than differences arising from label definitions or taxonomic annotation practices between datasets.

\subsection{Session construction and temporal framework}
Following dataset preparation and target-taxon harmonisation, a temporal framework was required to transform image-level phytoplankton observations into ecologically interpretable signals. Because the available IFCB observations were collected during discrete sampling periods separated by irregular temporal gaps, MedPlanktonSet was not treated as a continuous time series. Instead, temporal units were constructed from the observed sampling structure to preserve the temporal information available in the dataset while avoiding assumptions of continuous monitoring.

A recording session was defined as all images acquired at the same sampling location on the same recording date. The resulting temporal framework is illustrated in Figure 2. This definition linked individual image observations to the finest reliable temporal unit available in the dataset while preserving location-specific sampling histories. For each session, the first and last image timestamps were used to determine recording duration, and the total number of images was recorded.

Not all recording sessions were considered equally suitable for temporal analysis. Sessions containing very few observations or covering only a short recording period can produce unstable estimates of target occurrence because individual classifications exert a disproportionately large influence on the resulting signal. To reduce this effect, sessions were required to contain at least 20 images and span a minimum of 30 minutes between the first and last recorded observation. These thresholds were introduced as quality-control criteria following exploratory examination of the session distribution and were intended to exclude sparsely sampled recording periods while retaining the majority of available observations.

Reference temporal signals were constructed by aggregating expert-labelled image observations within each session. These signals should not be interpreted as expert-labelled bloom events. Rather, they represent session-level summaries derived from expert taxonomic annotations. For each session \(s\), a target-occurrence rate (\(TOR_s\)) was calculated as

\begin{equation}
TOR_s = \frac{N_{\mathrm{target},s}}{N_{\mathrm{total},s}}
\label{eq:tor_session}
\end{equation}

where \(N_{\mathrm{target},s}\) denotes the number of images assigned to the selected target taxa within session \(s\), and \(N_{\mathrm{total},s}\) denotes the total number of images recorded in that session. The same aggregation procedure was subsequently applied to AI-generated classifications, allowing reference and AI-derived temporal signals to be compared directly at equivalent temporal scales.

Session-level signals describe individual sampling events but may be sensitive to short-term variability arising from sampling intensity, local environmental conditions, or classification uncertainty. To investigate whether ecologically meaningful patterns persisted across repeated sampling occasions, session-level observations were further aggregated into fixed-length sampling-session windows. This approach allowed persistence and temporal trends to be examined despite the irregular spacing of observations in calendar time.

\begin{figure*}
\centering
\includegraphics[width=1\textwidth]{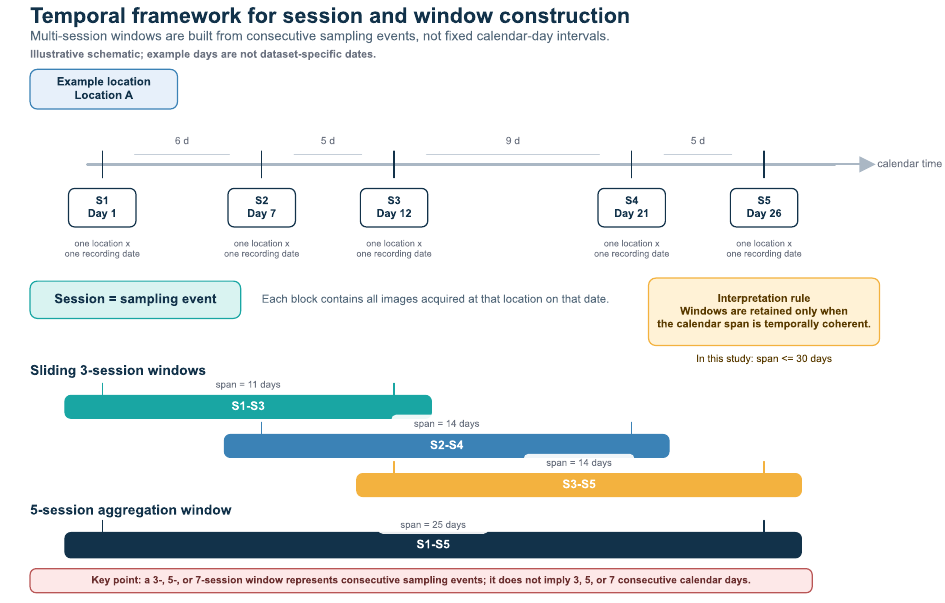}
\caption{Temporal framework used to aggregate image-level phytoplankton observations into monitoring-oriented signals.}
\label{fig:2}
\end{figure*}

Windows were constructed as sliding sequences of consecutive sampling sessions from the same location. As shown in Figure \ref{fig:2}, window length refers to the number of consecutive sampling events rather than fixed calendar-day intervals. Three window lengths were evaluated: 3-session, 5-session, and 7-session windows. These windows should not be interpreted as 3-day, 5-day, or 7-day calendar windows because sampling sessions were not evenly spaced in time. Instead, they represent fixed-length aggregations across successive sampling events. The use of multiple window lengths enabled assessment of how temporal aggregation influenced the stability and consistency of the resulting ecological indicators. A summary of the temporal units generated under this framework is provided in Table \ref{tab:3}.

\begin{table*}[t]
\centering
\caption{Summary of the temporal aggregation framework.}
\label{tab:3}
\small
\begin{tabular}{p{2.5cm}|p{4.0cm}|r|r|p{2.6cm}|c}
\hline
\textbf{Temporal unit} &
\textbf{Definition} &
\textbf{Total} &
\textbf{Retained} &
\textbf{Criterion} &
\textbf{Span (days)} \\
\hline

Recording session &
Images collected at a single location on a single recording date &
76 &
67 &
$\geq$20 images and $\geq$30 min duration &
Single day \\
\hline

3-session window &
Three consecutive sessions from the same location &
64 &
37 &
Calendar span $\leq$30 days &
5--28 \\
\hline

5-session window &
Five consecutive sessions from the same location &
54 &
14 &
Calendar span $\leq$30 days &
12--28 \\
\hline

7-session window &
Seven consecutive sessions from the same location &
48 &
9 &
Calendar span $\leq$30 days &
20--29 \\
\hline

\end{tabular}
\end{table*}

Within each session and window, temporal indicators were derived to characterise target occurrence, dominance, persistence, and temporal change. Dominance was defined as the taxon contributing the largest proportion of target observations within a session or window. Persistence described the repeated occurrence of a target taxon across successive sampling sessions, whereas temporal change was assessed through variation in target-occurrence rates across the aggregation window. These indicators were intended to summarise ecological patterns that may not be evident from individual image classifications alone.

Because the available observations were irregularly distributed in time, an additional constraint was applied during event-proxy analysis. Aggregation windows were retained only when the calendar span between the first and last session did not exceed 30 days. This restriction was introduced to ensure that indicators of persistence and temporal change were derived from temporally coherent sampling periods rather than from windows spanning extended gaps between observations. Event proxies were therefore evaluated only within windows considered suitable for interpreting short-term ecological dynamics.

It should be noted that neither MedPlanktonSet nor the SMHI IFCB dataset was originally designed as a benchmark for forecasting, or continuous monitoring-system development. Although both datasets contain timestamped image observations, the available records are characterised by irregular sampling intervals and do not include independently verified bloom-event annotations. Consequently, the objective of the present study is not to develop or validate a bloom-detection system directly. Instead, the available temporal information was used to construct a conservative monitoring-oriented framework in which image-level classifications could be aggregated into session-level and multi-session indicators. This approach provides a means of evaluating whether AI-derived phytoplankton classifications retain ecologically meaningful temporal patterns despite the absence of dedicated monitoring or bloom-event datasets.

\subsection{Image classification framework}
Following construction of the temporal framework described in Section 2.3, a supervised image-classification model was developed to assign phytoplankton images to the selected target taxa. The resulting image-level predictions were subsequently used for the temporal analyses described in Section 2.5. The classification task was defined using the eight target taxa identified through the harmonisation process described in Section 2.2 together with a background OTHER category, resulting in a nine-class classification problem.
Model development was performed exclusively using MedPlanktonSet. To minimise temporal leakage, images were partitioned at the recording-session level rather than through random image-level sampling. All images originating from the same recording session were assigned to a single subset. A session-aware, class-coverage-constrained splitting procedure was used to generate train, validation, and test partitions. Candidate session assignments were optimised to approximate 70:15:15 train–validation–test proportions while maintaining representation of all model classes across the three subsets. The optimisation procedure penalised class imbalance, image-count imbalance, deviations from the target split proportions, and the absence of any class from a subset. This approach ensured that model evaluation reflected generalisation across independent sampling events rather than repeated exposure to images acquired during the same recording session. The independent SMHI dataset was not used during training, validation, hyperparameter tuning, model selection, or checkpoint selection.

Prior to model training, all images were converted to RGB format and resized to a standardised input resolution of 96 × 96 pixels. This resolution was selected because it closely matched the scale of the available IFCB regions of interest (ROIs), avoided unnecessary image upsampling, and provided a computationally efficient input size for model development and comparison. To preserve biologically relevant morphology, the final preprocessing pipeline used aspect-ratio-preserving resizing with background padding rather than direct geometric stretching. Padding values were derived from the local image background to minimise artificial boundaries introduced during resizing. This strategy was intended to preserve morphological characteristics such as elongation, chain structure, circularity, and overall object shape that may otherwise be distorted by conventional resizing. The same deterministic preprocessing pipeline was applied to validation, test, and external-evaluation images.

Training images were augmented to improve robustness to orientation and minor acquisition variability. Augmentations included random horizontal flipping, random vertical flipping, and mild random rotation within ±20. Rotational augmentation was included because the orientation of an ROI within the image frame can vary between imaging systems, segmentation procedures, and sampling conditions. Augmentations were applied only during training. Validation, test, and external-evaluation images were not randomly augmented, ensuring deterministic and directly comparable evaluation. 

Several widely used convolutional neural network (CNN) architectures were evaluated during model development, including MobileNetV3-Small \citep{36}, ResNet-18 \citep{37}, ResNet-50 \citep{37}, EfficientNet-B0 \citep{38}, and DenseNet-121 \citep{39}. Table \ref{tab:4} presents the details and purpose of the selection of these models. These architectures were selected to represent a range of computational complexity and feature-extraction strategies relevant to automated phytoplankton monitoring. MobileNetV3-Small, EfficientNet-B0, ResNet-18, and DenseNet-121 were chosen because they provide comparatively efficient inference and are compatible with potential real-time or near-real-time monitoring applications. ResNet-50 was additionally included as a higher-capacity reference architecture to examine whether increased model complexity provided measurable improvements in classification performance. The final classification layer was modified to match the nine output classes. Models were trained using cross-entropy loss and the AdamW optimiser with a learning rate of 0.001, weight decay of 0.0001, and batch size of 64 for a maximum of 50 epochs on a CUDA-enabled GPU.
Class imbalance was addressed using square-root inverse-frequency weighting within the loss function. This approach reduced the influence of highly abundant classes while avoiding the instability that can arise when full inverse-frequency weighting is applied to rare classes. In addition, the OTHER category was downsampled to a maximum of 10,000 images in the training set to reduce background-class dominance during optimisation while retaining substantial representation of non-target observations. No downsampling was applied to the validation or test sets, ensuring that evaluation reflected the natural class distribution present in the held-out data.

\begin{table*}[t]
\centering
\caption{Convolutional neural network architectures comparison and selection choice}
\label{tab:4}
\small
\begin{tabular}{p{2.8cm}|c|c|p{4.2cm}}
\hline
\textbf{Model} &
\textbf{Parameters} &
\textbf{Deployment suitability} &
\textbf{Purpose} \\
\hline

MobileNetV3-Small &
2.5 M &
High &
Lightweight baseline representing deployment-oriented and real-time monitoring scenarios \\
\hline

ResNet-18 &
11.7 M &
Moderate &
Standard residual CNN baseline for comparison with lightweight architectures \\
\hline

ResNet-50 &
25.6 M &
Lower &
Higher-capacity residual architecture used as a reference model \\
\hline

EfficientNet-B0 &
5.3 M &
High &
Accuracy-efficiency architecture designed to balance predictive performance and computational cost \\
\hline

DenseNet-121 &
8.0 M &
Moderate &
Feature-reuse architecture included to evaluate densely connected representations \\
\hline

\end{tabular}
\end{table*}

\subsection{Temporal indicator construction and event-proxy framework}
The image-classification framework described in Section~2.4 generated predictions at the individual-image level. However, the primary objective of this study was not to evaluate image classifications in isolation, but to determine whether these classifications could be transformed into temporally meaningful indicators of phytoplankton occurrence. Consequently, expert-labelled observations and AI-generated predictions were aggregated using the common session and window framework described in Section~2.3. Applying the same aggregation procedure to both data sources ensured that reference and AI-derived indicators were constructed at equivalent temporal scales.

\subsubsection{Session-level temporal indicators}
For each recording session, the total number of images, the number of images assigned to the selected target taxa, and the number of images assigned to each target class were calculated separately for expert labels and AI predictions. The primary session-level indicator was the target-occurrence rate (\(TOR\)), as defined in Eq. \ref{eq:tor_session}. Taxon-specific occurrence rates were additionally calculated as

\begin{equation}
TOR_{c,s} = \frac{N_{c,s}}{N_{\mathrm{total},s}}
\label{eq:tor_taxon}
\end{equation}

where \(N_{c,s}\) represents the number of images assigned to taxon \(c\) within session \(s\).

The dominant taxon was defined as the target class contributing the largest number of observations within a session. Together, these indicators provided a description of overall target occurrence, taxonomic composition, and relative dominance at the scale of individual sampling events.

\subsubsection{Window-level temporal indicators}
Individual recording sessions represent discrete sampling events and may therefore be influenced by short-term variability arising from sampling intensity, environmental conditions, or classification uncertainty. To investigate whether ecologically meaningful patterns persisted across repeated sampling occasions, session-level indicators were further aggregated into fixed-length sampling-session windows.

As described in Section~2.3, windows were constructed as sliding sequences of consecutive sampling sessions from the same location using 3-, 5-, and 7-session window lengths. Because sampling intervals were irregular, these windows should not be interpreted as fixed calendar periods. Instead, they represent aggregation across successive sampling events.

For each window, a weighted target-occurrence rate was calculated as

\begin{equation}
TOR_w =
\frac{\sum_{i=1}^{w} N_{\mathrm{target},i}}
{\sum_{i=1}^{w} N_{\mathrm{total},i}}
\label{eq:tor_window}
\end{equation}

where \(w\) denotes the window length and \(i\) denotes the session index within the window. This formulation accounts for differences in the number of images acquired during individual sessions.

An unweighted mean target-occurrence rate was also calculated as

\begin{equation}
\overline{TOR}_w =
\frac{1}{w}
\sum_{i=1}^{w}
TOR_i
\label{eq:tor_mean}
\end{equation}

to provide a complementary summary of target occurrence across the sessions comprising a window.

Temporal change within each window was represented by a linear trend estimated across ordered sessions:

\begin{equation}
TOR_i = \beta_0 + \beta_1 S_i
\label{eq:tor_trend}
\end{equation}

where \(S_i\) denotes the session order within the window and \(\beta_1\) represents the trend coefficient. Positive values of \(\beta_1\) indicate increasing target occurrence across the window, whereas negative values indicate decreasing occurrence.

\subsubsection{Event-proxy construction}

The available datasets contain expert-labelled image observations and timestamps but do not include independently verified bloom-event annotations. Consequently, the study does not attempt to identify confirmed bloom events. Instead, threshold-based event proxies were constructed to evaluate whether AI-derived temporal indicators reproduced elevated, persistent, or increasing target-occurrence patterns observed in the expert-labelled reference data.

To account for differences in baseline target occurrence among sampling locations, location-specific thresholds were estimated from the distribution of reference target-occurrence rates. Elevated occurrence was defined using the 75th percentile of the reference distribution, whereas strong occurrence was defined using the 90th percentile. A minimum threshold of 0.01 was applied to avoid unrealistically small thresholds in locations exhibiting very low target occurrence.

Thresholds were estimated exclusively from expert-labelled calibration sessions derived from the training and validation partitions. The resulting thresholds were subsequently applied to the held-out test sessions and corresponding AI-derived indicators.

At the session level, three event-proxy categories were defined:

\begin{itemize}
\item \textbf{Elevated occurrence}: \(TOR_s\) greater than or equal to the location-specific elevated threshold.
\item \textbf{Strong occurrence}: \(TOR_s\) greater than or equal to the location-specific strong threshold.
\item \textbf{Persistent elevated occurrence}: an elevated session that is immediately preceded or followed by another elevated session from the same location in the ordered session sequence.
\end{itemize}

At the window level, four event-proxy categories were defined:

\begin{itemize}
\item \textbf{Elevated occurrence}: \(TOR_w\) greater than or equal to the elevated threshold.
\item \textbf{Strong occurrence}: \(TOR_w\) greater than or equal to the strong threshold.
\item \textbf{Increasing occurrence}: \(\beta_1 > 0\).
\item \textbf{Elevated-and-increasing occurrence}: simultaneous satisfaction of the elevated-occurrence and increasing-occurrence criteria.
\end{itemize}

These event proxies were intended to represent elevated, persistent, or increasing target-taxon occurrence patterns that may be relevant for monitoring applications, while avoiding unsupported claims of direct bloom detection.

\subsection{Cross-domain evaluation and statistical analysis}
The image-classification framework and temporal-indicator construction procedures described in Sections 2.4 and 2.5 were evaluated at three complementary levels: (i) image-level classification performance, (ii) temporal-indicator reconstruction, and (iii) event-proxy agreement. These analyses were designed to assess not only the ability of the model to classify individual phytoplankton images, but also its ability to reproduce monitoring-relevant temporal patterns after aggregation.

For each evaluated CNN architecture, the model checkpoint achieving the highest validation macro F1-score was retained for subsequent evaluation. Macro F1-score was used for checkpoint selection because it gives equal importance to all classes and is therefore more appropriate than overall accuracy for imbalanced phytoplankton datasets. Final image-level performance was subsequently evaluated using the held-out MedPlanktonSet test sessions only. Test data were not used during model training, hyperparameter optimisation, checkpoint selection, threshold calibration, or preprocessing design.
Image-level classification performance was assessed using accuracy, macro F1-score, weighted F1-score, and per-class precision, recall, and F1-score. Accuracy provided an overall measure of classification correctness, whereas macro F1-score evaluated performance equally across taxa regardless of class frequency. Weighted F1-score was additionally reported to characterise performance under the natural class distribution of the held-out dataset. 

Temporal indicators generated using the procedures described in Section 2.5 were evaluated by comparing AI-derived and reference target-occurrence signals at both session and window levels. Agreement between reference and AI-derived target-occurrence rates was assessed using mean absolute error (MAE), Pearson correlation, and Spearman rank correlation. MAE quantified the average magnitude of disagreement between reference and predicted occurrence rates. Pearson correlation quantified linear agreement between the two signals, whereas Spearman correlation assessed whether temporal units exhibiting higher target occurrence in the reference signal also tended to exhibit higher target occurrence in the AI-derived signal. Dominant-taxon agreement was additionally calculated as the proportion of temporal units for which the dominant AI-predicted taxon matched the dominant taxon identified from the reference observations.

Event-proxy performance was evaluated on held-out test sessions and windows using thresholds estimated exclusively from the training and validation partitions. For each event-proxy category, AI-derived event labels were compared with reference event labels constructed from expert-labelled image aggregations. Event agreement rate was defined as the proportion of temporal units for which the AI-derived and reference event-proxy labels were identical. Performance was further summarised using precision, recall, and F1-score together with counts of false-positive and false-negative event detections. This analysis was intended to determine whether AI-derived temporal indicators could reproduce elevated, persistent, or increasing target-occurrence patterns observed in the reference data, rather than to validate confirmed bloom events.

Cross-domain generalisation was assessed using the independent SMHI IFCB dataset. No SMHI images were used during model training, validation, hyperparameter optimisation, checkpoint selection, temporal aggregation, or threshold calibration. Because the two datasets differed in taxonomic annotation structure, external evaluation was restricted to taxa exhibiting exact label correspondence between MedPlanktonSet and SMHI. Classification performance on SMHI was therefore interpreted as a strict assessment of transferability under geographic, temporal, imaging, taxonomic, and annotation-domain shift.
All analyses were implemented in Python using PyTorch for model training and inference, and pandas, NumPy, SciPy, scikit-learn, and Matplotlib for data processing, statistical analysis, and visualisation. Random seeds were fixed where applicable to improve reproducibility. Model configurations, class mappings, image-level predictions, session-level summaries, window-level summaries, and evaluation outputs were archived to support reproducibility and downstream analysis.

\section{Results}
\subsection{Analytical dataset characteristics}
Application of the target-selection and temporal-filtering procedures resulted in a large but substantially constrained analytical dataset. The final MedPlanktonSet target framework contained 20,273 images belonging to the eight selected target taxa (Table \ref{tab:2}), representing 26.2\% of the 77,271 available images. The remaining 56,998 images (73.8\%) were assigned to the OTHER category and retained as part of the background image stream. Consequently, target occurrence was evaluated relative to the full phytoplankton image record rather than within a pre-filtered subset of target taxa, providing a more realistic basis for assessing monitoring-oriented temporal indicators.

\begin{figure}
\centering
\includegraphics[width=1\textwidth]{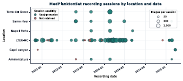}
\caption{Temporal and spatial distribution of MedPlanktonSet recording sessions used in the analysis.}
\label{fig:3}
\end{figure}

The relationship between recording sessions and multi-session aggregation windows is illustrated in Figure \ref{fig:2}. Of the 76 recording sessions identified in MedPlanktonSet, 67 satisfied the minimum image-count and recording-duration criteria and were retained for temporal analysis, corresponding to a session-retention rate of 88.2\%. This indicates that the majority of recording events contained sufficient observations to support estimation of session-level target-occurrence rates. The retained and excluded sessions, together with their temporal and spatial distribution across sampling locations, are shown in Figure \ref{fig:3}.

The temporal constraints became more restrictive at the multi-session window level. Application of the maximum 30-day calendar-span criterion retained 37 of 64 possible 3-session windows (57.8\%), 14 of 54 possible 5-session windows (25.9\%), and 9 of 48 possible 7-session windows (18.8\%). As illustrated by the uneven temporal distribution of recording sessions across locations (Fig. 3), increasing window length reduced the number of temporally coherent aggregation units available for event-proxy analysis. Thus, although longer windows provided greater temporal smoothing and improved the assessment of persistence and short-term trends, they also substantially reduced the number of independent temporal units available for validation. This trade-off between temporal stability and sample availability became increasingly pronounced as window length increased.

The external-domain evaluation was necessarily narrower than the MedPlanktonSet temporal framework. Although eight taxa were retained within the MedPlanktonSet target framework, only four taxa exhibited exact label correspondence and usable image records within the SMHI dataset (Table \ref{tab:2}). Consequently, cross-domain evaluation was intentionally restricted to these exact taxonomic matches, reducing the likelihood that observed performance differences could be attributed to label-definition inconsistencies rather than genuine geographic, imaging, taxonomic, or annotation-domain shift. The resulting SMHI evaluation therefore represents a conservative assessment of external transferability.

\subsection{Classification model development and transferability analysis}
\subsubsection{Input representation and class-framework analysis}
As the objective of this study is to evaluate monitoring-oriented phytoplankton classification under realistic deployment conditions, input-representation decisions were assessed using both in-domain performance on MedPlanktonSet and cross-domain transferability to the independent SMHI dataset. Consequently, preprocessing strategies were not selected solely on the basis of in-domain classification accuracy but also on their ability to generalise across independently acquired IFCB image collections.
Initial diagnostic experiments were performed using MobileNetV3-Small, a lightweight convolutional neural network selected as a computationally efficient baseline representative of potential real-time monitoring applications. These experiments were designed to evaluate preprocessing and class-framework choices before comparing alternative CNN architectures. Evaluation was initially conducted using a closed-set eight-taxon classification task to determine whether the selected target taxa could be learned reliably prior to construction of the full monitoring framework.

The results revealed a clear trade-off between in-domain classification performance and external-domain transferability (Table \ref{tab:5}). Direct resizing to 96 × 96 pixels produced the strongest in-domain MedPlanktonSet performance, achieving a test accuracy of 0.9645 and a macro F1-score of 0.9452. However, when evaluated on the strict exact-label SMHI dataset, performance deteriorated substantially, with an accuracy of only 0.0966 and a supported-label macro F1-score of 0.2801. These results indicate that strong in-domain classification performance alone did not guarantee robust transfer to independently acquired IFCB imagery. This finding is particularly relevant for monitoring applications, where models may ultimately be deployed on image streams collected under different environmental, imaging, or annotation conditions than those used during training.
Aspect-ratio-preserving border-median padding was therefore evaluated as an alternative preprocessing strategy designed to preserve the original geometric proportions of phytoplankton regions of interest. Although aspect-preserving preprocessing at 96 × 96 pixels produced slightly lower MedPlanktonSet performance than direct resizing, achieving a test accuracy of 0.9455 and a macro F1-score of 0.8905, it substantially improved external-domain performance, increasing SMHI accuracy to 0.6636 and supported-label macro F1-score to 0.5878. This improvement suggests that preserving the original geometric structure of phytoplankton regions of interest reduced sensitivity to dataset-specific image characteristics and improved generalisation across independently acquired image collections.

\begin{table*}[t]
\centering
\caption{Diagnostic experiments used to define the final monitoring-oriented preprocessing framework. All diagnostic experiments were performed using MobileNetV3-Small.}
\label{tab:5}
\scriptsize
\begin{tabular}{p{2.0cm}|p{0.6cm}|p{1.8cm}|p{1.8cm}|p{1.0cm}|p{1.0cm}|p{1.0cm}|p{1.0cm}|p{2.0cm}}
\hline
\textbf{Input representation} &
\textbf{Size} &
\textbf{Augmentation} &
\textbf{Intensity} &
\textbf{Med Acc.} &
\textbf{Med F1} &
\textbf{SMHI Acc.} &
\textbf{SMHI F1} &
\textbf{Interpretation} \\
\hline

Direct resize &
96$\times$96 &
None &
RGB &
\underline{0.9645} &
\textbf{0.9452} &
0.0966 &
0.2801 &
Best Med in-domain result but poor external transfer \\
\hline

Aspect-ratio padding, border-median fill &
96$\times$96 &
None &
RGB &
0.9455 &
0.8905 &
\underline{0.6636} &
\underline{0.5878} &
Selected input representation for transfer-oriented framework \\
\hline

Aspect-ratio padding, border-median fill &
128$\times$128 &
None &
RGB &
\textbf{0.9659} &
\underline{0.9120} &
0.5757 &
0.4801 &
Improved Med performance but reduced SMHI transfer relative to 96$\times$96 aspect padding \\
\hline

Aspect-ratio padding, border-median fill &
96$\times$96 &
Cardinal rotation &
RGB &
0.9511 &
0.8928 &
0.6434 &
0.4901 &
Did not improve external transfer \\
\hline

Aspect-ratio padding, border-median fill &
96$\times$96 &
Arbitrary rotation &
RGB &
0.9352 &
0.8235 &
0.0465 &
0.2056 &
Rejected; harmful to external transfer \\
\hline

Aspect-ratio padding, border-median fill &
96$\times$96 &
Mild rotation ($\pm20^\circ$) &
Gray standardisation &
0.9373 &
0.8626 &
0.3225 &
0.5404 &
Not retained; did not improve overall transfer \\
\hline

Aspect-ratio padding, border-median fill &
96$\times$96 &
Mild rotation ($\pm20^\circ$) &
Gray equalisation &
0.9235 &
0.8487 &
0.3240 &
0.5270 &
Not retained; did not improve overall transfer \\
\hline

Aspect-ratio padding, border-median fill &
96$\times$96 &
Mild rotation ($\pm20^\circ$) &
RGB &
0.9168 &
0.7931 &
\textbf{0.6778} &
\textbf{0.6128} &
Final selected monitoring-oriented configuration \\
\hline

\end{tabular}
\end{table*}

Input size also influenced the balance between in-domain and external-domain performance. Increasing the aspect-preserved input size from 96 × 96 to 128 × 128 pixels improved MedPlanktonSet macro F1-score from 0.8905 to 0.9120. However, SMHI performance decreased, with accuracy falling from 0.6636 to 0.5757 and supported-label macro F1-score declining from 0.5878 to 0.4801. Therefore, although the larger input size improved in-domain classification performance, it reduced external-domain transferability and did not provide the most balanced performance across datasets. Given the study's emphasis on monitoring under domain shift, aspect-ratio-preserving border-median padding at 96 × 96 pixels was retained as the preferred input representation.

Additional augmentation and intensity-processing experiments were subsequently performed using the 96 × 96 aspect-preserved representation (Table \ref{tab:5}). Cardinal rotation produced performance comparable to the baseline aspect-preserved configuration but did not improve external-domain transfer. In contrast, arbitrary rotation substantially reduced both in-domain and external-domain performance, suggesting that excessive rotational augmentation was detrimental for the available datasets. Gray-level standardisation and gray equalisation also failed to improve transferability and generally reduced classification performance relative to the baseline RGB representation. These results indicate that aggressive augmentation and intensity transformations offered little benefit for cross-domain generalisation. Consequently, the final preprocessing pipeline retained RGB imagery, aspect-ratio-preserving border-median padding, and a conservative mild rotational augmentation strategy (±20°) during training.

The final methodological decision concerned the classification framework itself. Although the closed-set target-only experiments demonstrated that the selected taxa could be classified and transferred with reasonable accuracy, such models are unsuitable for monitoring because they force every image into one of the selected target classes. Temporal target-occurrence rates, however, must be calculated relative to the complete phytoplankton image stream. The classification framework was therefore expanded from the closed-set eight-taxon task to a monitoring-oriented nine-class task comprising the eight selected target taxa and an OTHER category representing non-target observations. This design enabled target-occurrence rates to be estimated relative to the full image stream and provided the denominator required for the temporal-indicator and event-proxy analyses that form the core monitoring framework of this study.

\subsubsection{CNN architecture comparison under monitoring framework}
Following selection of the input representation and monitoring-oriented class framework, candidate CNN architectures were evaluated using the final nine-class classification task comprising the eight target taxa and the OTHER category. All models were trained using identical session-aware train, validation, and test partitions, the same aspect-ratio-preserving 96 × 96 preprocessing pipeline, RGB input, mild rotational augmentation, square-root inverse-frequency class weighting, and a maximum of 10,000 OTHER images in the training set. Consequently, observed performance differences can be attributed primarily to architectural differences rather than variations in data partitioning, preprocessing, or class balance.

Model comparison was not based solely on overall classification accuracy because the downstream temporal analyses depend on accurate estimation of target-occurrence rates. In addition to standard multi-class metrics, performance was therefore evaluated using TARGET-versus-OTHER precision, recall, and F1-score. For these metrics, the eight selected target taxa were collapsed into a single TARGET category and compared against the OTHER category representing non-target observations. These metrics assess the model's ability to distinguish monitoring-relevant target observations from the broader non-target image stream and are therefore directly related to the reliability of subsequent temporal-indicator calculations.

\begin{figure}
\centering
\includegraphics[width=0.8\textwidth]{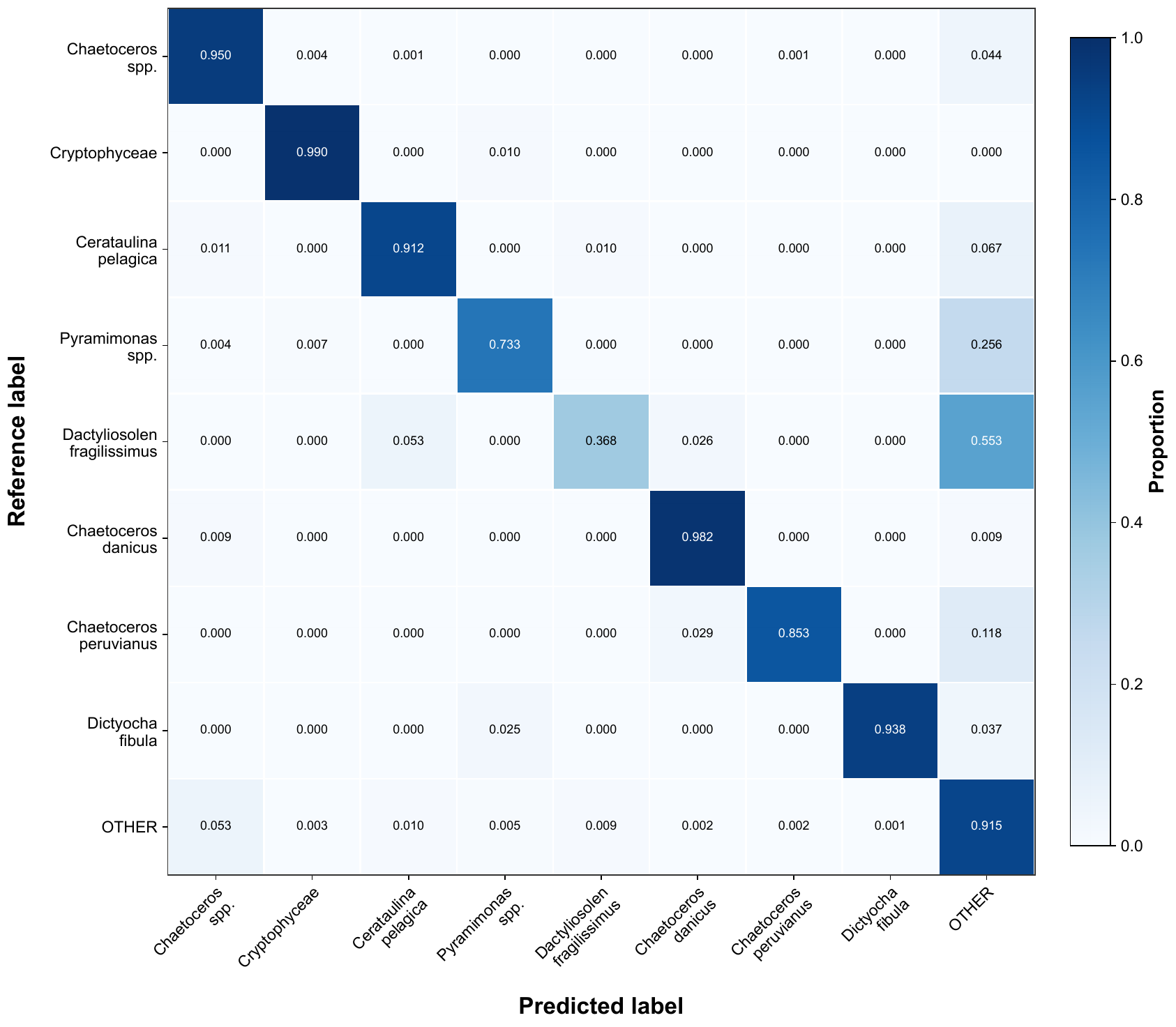}
\caption{Confusion matrix of the final EfficientNet-B0 model evaluated on held-out MedPlanktonSet test sessions.}
\label{fig:4}
\end{figure}

The architecture comparison revealed that EfficientNet-B0 and ResNet-18 provided the strongest overall performance under the monitoring-oriented framework (Table \ref{tab:6}). ResNet-18 achieved the highest multi-class macro F1-score (0.8026), marginally exceeding EfficientNet-B0 (0.7931). However, EfficientNet-B0 achieved the highest TARGET F1-score (0.8575), with a TARGET precision of 0.7939 and a TARGET recall of 0.9321. ResNet-18 produced a very similar TARGET F1-score (0.8540) but with lower TARGET recall (0.9073). DenseNet-121 also demonstrated strong performance, achieving a TARGET F1-score of 0.8487, although its macro F1-score remained lower than that of EfficientNet-B0 and ResNet-18. MobileNetV3-Small and ResNet-50 produced lower TARGET-versus-OTHER discrimination and lower macro F1-scores than the leading architectures, although MobileNetV3-Small remained competitive given its substantially lower computational complexity. Figure \ref{fig:4} presents confusion matrix for the selected EfficientNet-B0 monitoring model.

\begin{table*}[t]
\centering
\caption{Comparison of CNN architectures under the monitoring-oriented nine-class framework comprising the eight selected target taxa and the OTHER category.}
\label{tab:6}
\small
\begin{tabular}{p{2.5cm}|p{1.5cm}|p{1.5cm}|p{1.5cm}|p{1.5cm}|p{1.5cm}|p{1.5cm}}
\hline
\textbf{Model} &
\textbf{Accuracy} &
\textbf{Macro F1} &
\textbf{Weighted F1} &
\textbf{TARGET Precision} &
\textbf{TARGET Recall} &
\textbf{TARGET F1} \\
\hline

EfficientNet-B0 &
\textbf{0.9168} &
\underline{0.7931} &
\textbf{0.9201} &
\underline{0.7939} &
\textbf{0.9321} &
\textbf{0.8575} \\
\hline

ResNet-18 &
\underline{0.9156} &
\textbf{0.8026} &
\underline{0.9181} &
\textbf{0.8066} &
\underline{0.9073} &
\underline{0.8540} \\
\hline

DenseNet-121 &
0.9138 &
0.7633 &
0.9152 &
0.7999 &
0.9038 &
0.8487 \\
\hline

MobileNetV3-Small &
0.8912 &
0.7381 &
0.8907 &
0.7511 &
0.9052 &
0.8210 \\
\hline

ResNet-50 &
0.8679 &
0.6632 &
0.8716 &
0.7109 &
0.8697 &
0.7823 \\
\hline

\end{tabular}
\end{table*}

Several observations emerge from these results. First, overall classification accuracy alone was not sufficient to characterise monitoring suitability, as models with similar accuracy values exhibited different abilities to separate target observations from the background image stream. Second, the strongest-performing architectures consistently achieved high TARGET recall (>0.90), indicating that most target observations were successfully retained during classification. This is particularly important for temporal monitoring because missed target observations directly reduce estimated target-occurrence rates and may weaken reconstruction of temporal signals. Third, the relatively small performance differences between EfficientNet-B0, ResNet-18, and DenseNet-121 suggest that the monitoring framework itself contributed substantially to overall performance, whereas increasing architectural complexity beyond this level did not produce proportional improvements.

Model selection was guided primarily by monitoring-oriented performance rather than by image-level classification metrics alone. Although ResNet-18 achieved the highest macro F1-score (0.8026), EfficientNet-B0 provided the strongest TARGET-versus-OTHER discrimination, achieving the highest TARGET recall (0.9321) and TARGET F1-score (0.8575) while maintaining comparable overall accuracy, weighted F1-score, and macro F1-score. Because the downstream temporal indicators depend on accurate estimation of target occurrence within the complete image stream, minimising missed target observations was considered particularly important. EfficientNet-B0 was therefore selected for downstream temporal-indicator and event-proxy analyses, as it provided the most suitable balance between multi-class classification performance, TARGET-versus-OTHER discrimination, and computational efficiency. This selection criterion reflects the central objective of the study, namely the reliable reconstruction of monitoring-relevant temporal signals rather than the maximisation of image-level classification accuracy alone.

\subsection{Temporal signal reconstruction and event-proxy performance}
The central objective of this study was to determine whether AI-derived phytoplankton classifications could be transformed into monitoring-relevant temporal indicators rather than merely achieving high image-level classification performance. To evaluate this question, image-level predictions generated by the final EfficientNet-B0 model were aggregated using the session and window framework described in Section 2.5 and compared directly with reference temporal signals derived from expert-labelled observations.
At the session level, AI-derived target-occurrence rates closely reproduced the reference signals constructed from expert-labelled observations (Table \ref{tab:7a}). Across all 76 recording sessions, the mean absolute error between reference and AI-derived target-occurrence rates was 0.0432, while Pearson and Spearman correlations reached 0.9696 and 0.9568, respectively. These results demonstrate that temporal aggregation substantially reduced the influence of image-level classification errors and enabled reconstruction of temporal signals that closely approximated those derived from expert-labelled observations. Importantly, the strong correlations indicate that AI-derived classifications preserved both the magnitude and temporal ordering of target occurrence through time, directly addressing the primary objective of the study.

\begin{table}[t]
\centering
\caption{Agreement between AI-derived and expert-derived target-occurrence signals at different temporal aggregation levels.}
\label{tab:7a}
\small
\begin{tabular}{l|c|c|c|c|c}
\hline
\textbf{Analysis level} &
\textbf{Total $n$} &
\textbf{Usable $n$} &
\textbf{Target-rate MAE} &
\textbf{Pearson $r$} &
\textbf{Spearman $\rho$} \\
\hline

Session level &
76 &
67 &
\textbf{0.0432} &
0.9696 &
0.9568 \\
\hline

3-session windows &
64 &
37 &
\underline{0.0471} &
0.9750 &
\underline{0.9828} \\
\hline

5-session windows &
54 &
14 &
0.0480 &
\underline{0.9826} &
0.9819 \\
\hline

7-session windows &
48 &
9 &
0.0490 &
\textbf{0.9851} &
\textbf{0.9832} \\
\hline

\end{tabular}
\end{table}

\begin{figure}
\centering
\includegraphics[width=0.96\textwidth]{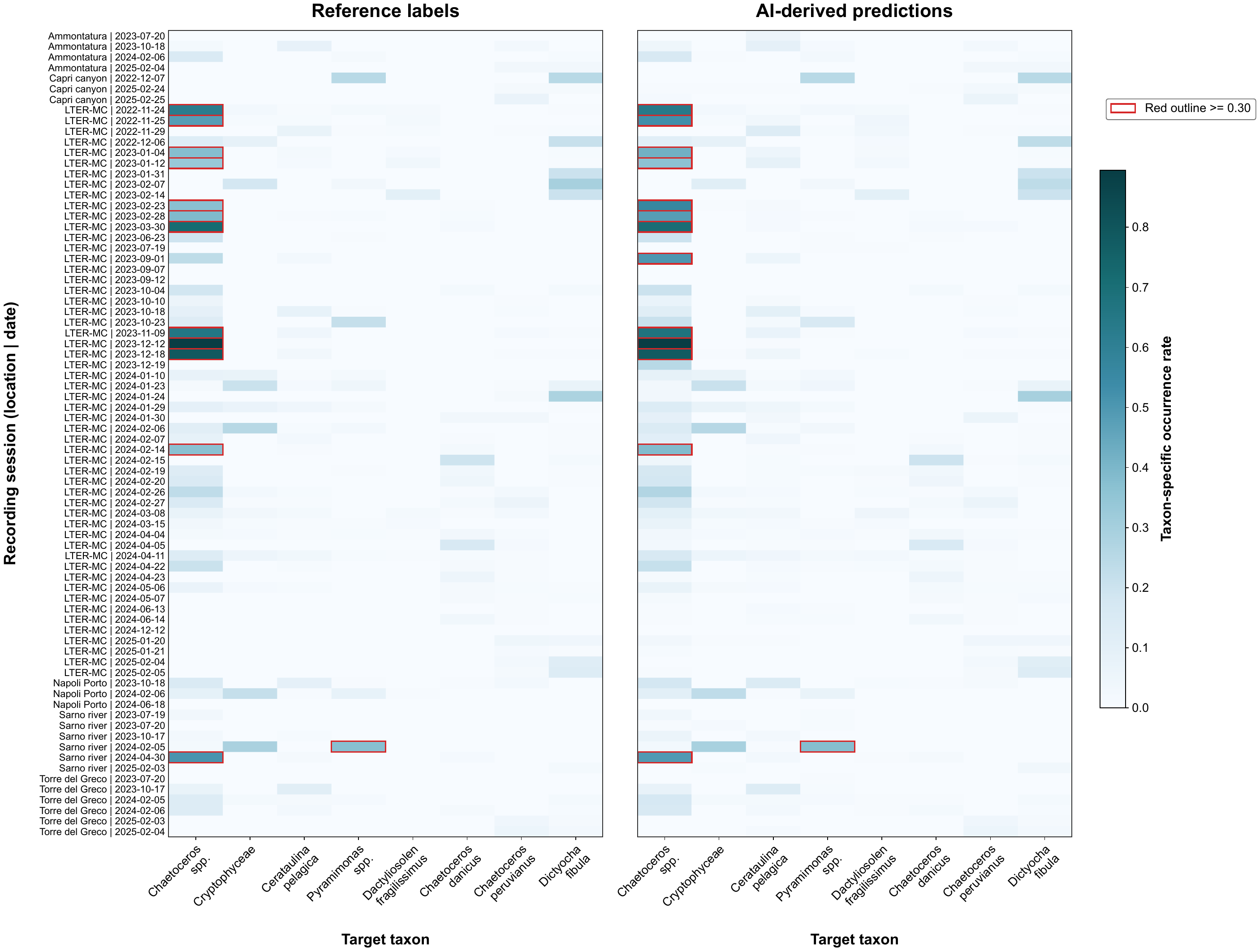}
\caption{Taxon-specific session-level occurrence rates derived from expert labels and AI predictions across all MedPlanktonSet recording sessions.}
\label{fig:5}
\end{figure}

Figure \ref{fig:5} compares the taxon-specific temporal signals obtained from expert labels and AI predictions across all recording sessions. Each cell represents the occurrence rate of a selected target taxon relative to the total number of images within a session. Cells outlined in white indicate taxon-specific occurrence rates greater than or equal to 0.30 and are included solely to facilitate visual identification of elevated taxon occurrence. This threshold was not used in any statistical analysis, event-proxy definition, or model evaluation. Although local discrepancies were evident for some taxa and sessions, the major occurrence hotspots, dominant taxa, and periods of elevated taxon occurrence were generally preserved. The close correspondence between the two heatmaps indicates that the aggregation process retained not only overall target occurrence but also much of the underlying taxonomic structure present in the expert-derived reference signal.

\begin{figure}
\centering
\includegraphics[width=0.9\textwidth]{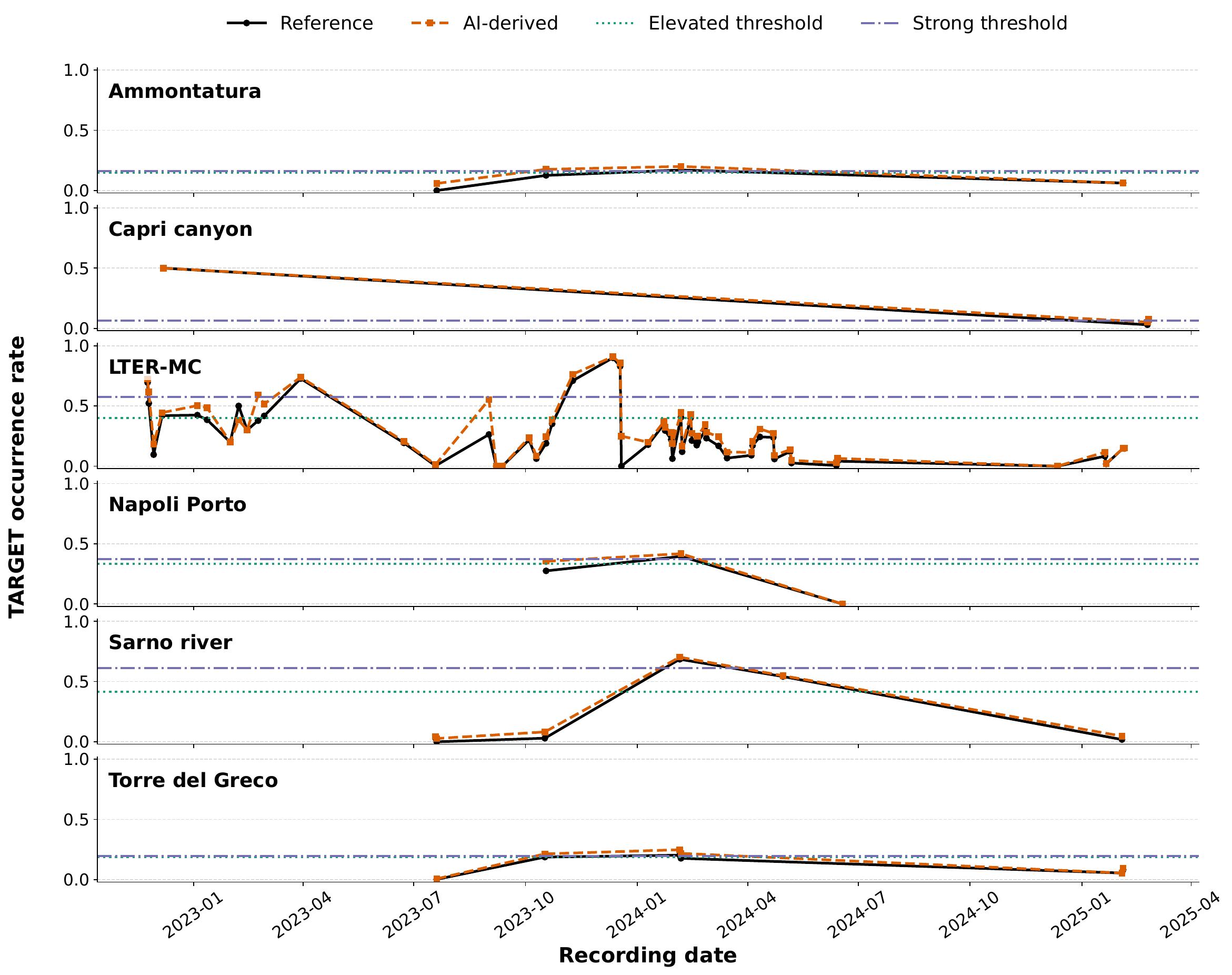}
\caption{Session-level TARGET occurrence rates across sampling locations.}
\label{fig:6}
\end{figure}

Figure \ref{fig:6} presents total target-occurrence rates after collapsing all selected taxa into a single TARGET category, where TARGET represents any observation belonging to one of the eight selected target taxa and OTHER represents all non-target observations. This representation reflects the primary monitoring objective of determining whether target taxa are increasing, decreasing, or elevated within the overall image stream. Across locations, AI-derived target-occurrence trajectories closely followed the expert-derived reference trajectories, reproducing both major peaks and periods of low occurrence. Importantly, the similarity between the two signals indicates that image-level classification errors did not substantially alter the temporal dynamics of target occurrence after aggregation. From a monitoring perspective, this finding is particularly significant because operational interpretation often depends more on accurate recovery of temporal trends than on perfect classification of every individual image.

\begin{figure}
\centering
\includegraphics[width=0.6\textwidth]{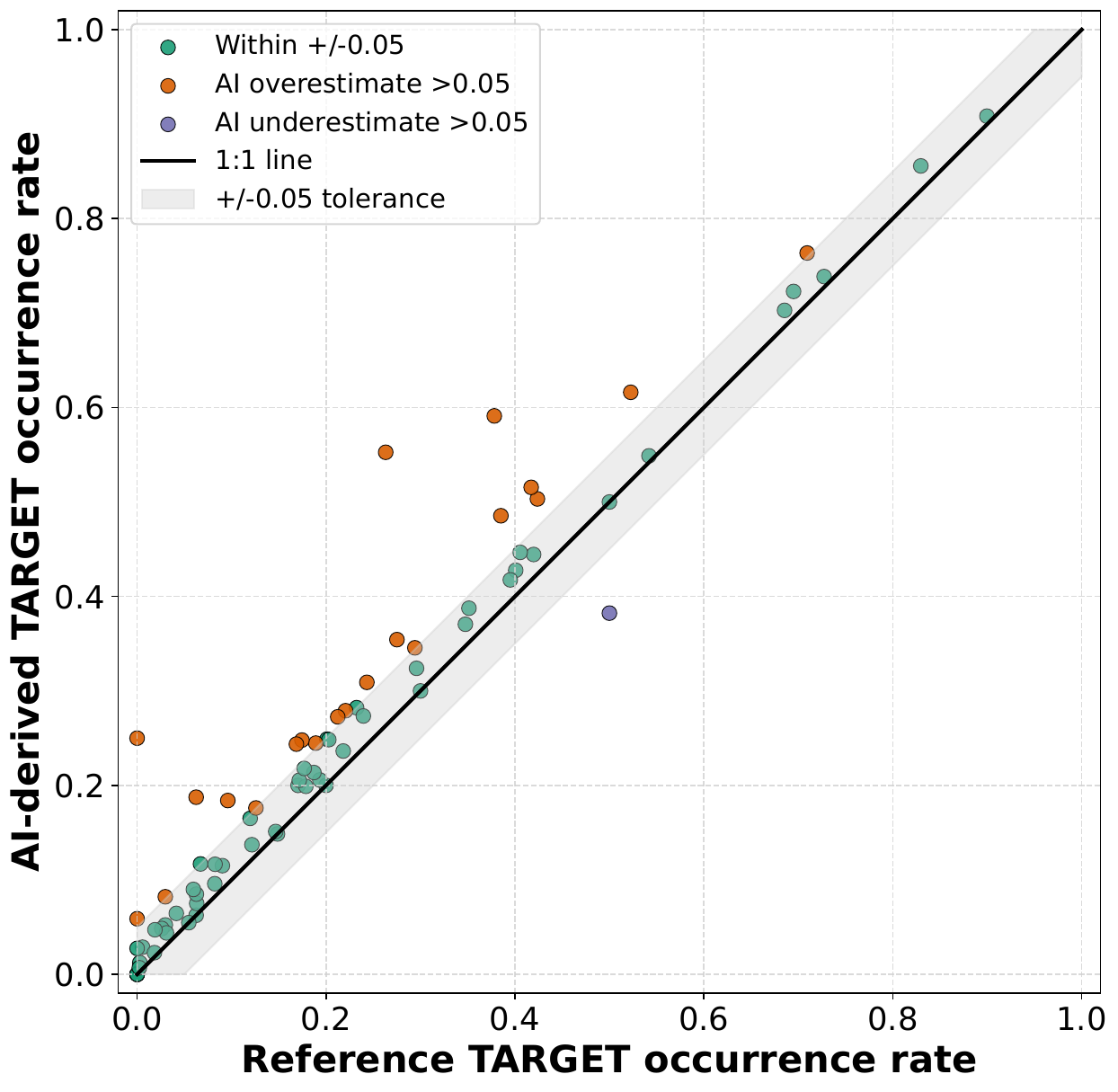}
\caption{Agreement between reference and AI-derived session-level TARGET occurrence rate.}
\label{fig:7}
\end{figure}

Figure \ref{fig:7} provides a direct comparison between expert-derived and AI-derived session-level target-occurrence rates. Most sessions lie close to the one-to-one relationship, consistent with the low mean absolute error and strong correlations reported in Table \ref{tab:7a}. Of the 76 sessions, 54 fell within the ±0.05 tolerance band. Deviations beyond this range were predominantly positive, with 21 sessions exhibiting AI overestimation greater than 0.05 compared with only a single session exhibiting underestimation greater than 0.05. Thus, residual errors were not symmetrically distributed, and the model showed a tendency to slightly overestimate target occurrence rather than miss substantial target-occurrence signals. Importantly, however, these deviations rarely altered the overall temporal trajectory, and the underlying temporal structure remained largely preserved.

Temporal aggregation was subsequently extended to multi-session windows to evaluate whether temporal smoothing further improved signal reconstruction. Strong agreement was maintained across all window lengths (Table \ref{tab:7a}). Three-session windows achieved a target-occurrence-rate MAE of 0.0471, Pearson correlation of 0.9750, and Spearman correlation of 0.9828. Five-session and seven-session windows produced similarly strong agreement, with Pearson correlations increasing to 0.9826 and 0.9851, respectively. Notably, agreement increased as aggregation length increased, indicating that residual image-level classification errors behaved primarily as short-term noise rather than systematic temporal bias. Consequently, temporal aggregation progressively enhanced signal stability while preserving the underlying temporal structure of target occurrence.

Longer windows, however, substantially reduced the number of usable temporal units available for analysis. Of the 64 possible three-session windows, 37 satisfied the temporal-coherence criteria, compared with 14 of 54 possible five-session windows and 9 of 48 possible seven-session windows. This reflects the irregular sampling structure of the dataset: longer windows provided greater temporal smoothing but were increasingly likely to span large calendar gaps and therefore became unsuitable for short-term event interpretation. Within the available dataset, three-session windows provided the most practical balance between temporal smoothing and analytical coverage, although this should not be interpreted as a universally optimal monitoring window.

Held-out event-proxy validation was performed using thresholds calibrated exclusively from training and validation sessions, ensuring that the test sessions remained independent of the calibration process. Results are summarised in Table \ref{tab:8}. For elevated-occurrence events, two reference events were present, and both were correctly identified by the AI-derived signal, with no false positives or false negatives. Similarly, the single strong-occurrence event and the single persistent elevated-occurrence event were both correctly reproduced. Although these results yielded perfect agreement within the held-out dataset, the number of event examples was extremely limited and therefore does not provide sufficient evidence to claim universally reliable event detection.

\begin{table}[t]
\centering
\caption{Held-out event-proxy validation comparing AI-derived and expert-derived temporal indicators.}
\label{tab:8}
\small
\begin{tabular}{p{2.2cm}|p{1.0cm}|p{1.0cm}|p{1.0cm}|p{1.0cm}|p{1.0cm}|p{1.8cm}|p{1.5cm}|p{1.3cm}}
\hline
\textbf{Event proxy} &
\textbf{Test sessions} &
\textbf{True events} &
\textbf{AI events} &
\textbf{False Positive} &
\textbf{False Negative} &
\textbf{Agreement} &
\textbf{Precision} &
\textbf{Recall / F1} \\
\hline

Elevated occurrence &
11 &
2 &
2 &
0 &
0 &
1.00 &
1.00 &
1.00 \\
\hline

Strong occurrence &
11 &
1 &
1 &
0 &
0 &
1.00 &
1.00 &
1.00 \\
\hline

Persistent elevated occurrence &
11 &
1 &
1 &
0 &
0 &
1.00 &
1.00 &
1.00 \\
\hline

\end{tabular}
\end{table}

Importantly, the event-proxy analysis does not constitute validation of confirmed bloom detection because independent bloom-event annotations were unavailable. Instead, it evaluates whether AI-derived classifications can reproduce threshold-based temporal indicators derived from expert-labelled observations. Within this context, the results suggest that AI-derived temporal signals were capable of recovering the limited number of elevated-occurrence patterns present in the held-out data.

The session-level, window-level, and event-proxy analyses provide a positive answer to the central question of this study. Although image-level classifications were not error-free, temporal aggregation substantially reduced the impact of those errors and enabled reconstruction of temporal signals that closely matched expert-derived reference signals. Importantly, the strongest evidence for the utility of the framework was not image-level classification accuracy itself, but its ability to recover temporal occurrence trajectories, dominant taxa, and elevated-occurrence patterns after aggregation. The principal contribution of the framework therefore lies in demonstrating that AI-derived phytoplankton classifications can be transformed into ecologically meaningful temporal indicators capable of supporting monitoring-oriented analysis of large IFCB image streams. Future validation on larger and more temporally continuous datasets containing independent bloom-event annotations will be required before such systems can be considered for operational bloom-monitoring applications. 

\section{Challenges, limitations, and implications for deployment}
The results demonstrate that AI-derived phytoplankton classifications can be transformed into monitoring-relevant temporal indicators; however, several limitations must be considered before such systems can be interpreted as operational bloom-detection tools. At the same time, the findings provide important insights into how AI may contribute to future phytoplankton monitoring workflows.

A key outcome of this study is that image-level classification performance alone was not the most informative measure of monitoring utility. Although classification errors were present, temporal aggregation substantially reduced their influence, allowing AI-derived signals to closely reproduce expert-derived temporal occurrence patterns. Strong agreement was observed at both the session level and the multi-session window level, indicating that many image-level errors behaved as local noise rather than systematic temporal distortion. This finding suggests that the value of AI classifiers for monitoring may lie less in achieving perfect image-level identification and more in their ability to recover ecologically meaningful temporal patterns from large image streams.

The first major limitation concerns the absence of independent bloom-event annotations. MedPlanktonSet provided expert-labelled images and timestamps but was not originally designed as a bloom-monitoring dataset. Consequently, the study evaluated elevated-occurrence proxies rather than confirmed bloom events. The strong agreement observed between AI-derived and expert-derived event proxies demonstrates the feasibility of the proposed framework, but the limited number of held-out events means that these results should not be interpreted as definitive evidence of operational bloom-detection capability. Validation against independently confirmed ecological events remains an important requirement for future work.

A second limitation arises from the temporal structure of the available data. Recording sessions were irregularly distributed across locations and time, and increasing the aggregation window length substantially reduced the number of temporally coherent units available for analysis. This highlights an important practical consideration for deployment: effective AI-based monitoring depends not only on classifier performance but also on the availability of sufficiently frequent and temporally consistent observations. Even highly accurate classifiers cannot reliably characterise ecological dynamics when observations are sparse or discontinuous.

Third, cross-domain transfer remained a significant challenge. The experiments demonstrated that preprocessing choices strongly influenced transferability, with aspect-ratio-preserving representations substantially outperforming direct image resizing on the independent SMHI dataset. Nevertheless, external-domain performance remained lower than in-domain performance, reflecting realistic differences in imaging conditions, ROI extraction, background appearance, morphology presentation, environmental context, and taxonomic annotation practices. These findings reinforce a broader challenge in ecological AI: strong performance within a single dataset does not necessarily translate into reliable performance across regions, instruments, or monitoring programmes. Future systems will therefore require broader multi-region training data, harmonised taxonomic frameworks, local calibration procedures, and continued external validation.

The findings also have implications for the design of future monitoring datasets. Many existing phytoplankton image collections have been developed primarily for classification research rather than temporal monitoring. As demonstrated in this study, timestamps and taxonomic labels alone are often insufficient for rigorous event-level validation. Monitoring-oriented datasets should therefore include consistent temporal sampling, harmonised metadata standards, sufficient image volumes per sampling period, and independent ecological annotations that allow bloom development, persistence, and decline to be evaluated directly. Such datasets would enable more robust assessment of whether AI-derived temporal indicators correspond to real ecological events.

Despite these limitations, the results support the future potential of AI-assisted phytoplankton monitoring. Under a monitoring-oriented data-collection framework, AI classifiers could provide near-real-time estimates of target-taxon occurrence, identify periods of elevated occurrence, prioritise expert review, and support the analysis of image streams that are impractical to inspect manually. Importantly, the present study demonstrates that AI-derived classifications can be transformed into temporal indicators that closely approximate expert-derived temporal signals. This suggests that future monitoring systems may benefit from treating AI classification not as an endpoint, but as the first stage of a broader temporal-analysis pipeline designed to support ecological interpretation and decision-making.

\section{Conclusion}
This study investigated whether AI-derived phytoplankton classifications can be transformed into monitoring-relevant temporal indicators rather than being evaluated solely through image-level classification accuracy. Using MedPlanktonSet, a session-aware classification and aggregation framework was developed to link IFCB image predictions with recording sessions, multi-session windows, and elevated-occurrence event proxies. The final monitoring-oriented model achieved strong separation between target taxa and non-target observations, and aggregation of image-level predictions produced temporal signals that closely matched reference signals derived from expert-labelled observations. High agreement was observed at both session and multi-session scales, indicating that temporal aggregation substantially reduced the impact of image-level classification errors and preserved monitoring-relevant occurrence patterns. The study also demonstrated that cross-domain transfer remains a major challenge. Although aspect-ratio-preserving preprocessing substantially improved transferability to the independent SMHI dataset, performance still declined outside the training domain, highlighting the need for broader training data, harmonised taxonomic frameworks, and continued external validation. Furthermore, the absence of independent bloom-event annotations limited evaluation to elevated-occurrence proxies rather than confirmed bloom events. The findings provide evidence that AI-derived phytoplankton classifications can be transformed into ecologically meaningful temporal indicators that closely reproduce expert-derived temporal signals. The principal contribution of the framework therefore lies not in achieving perfect image-level classification, but in enabling large IFCB image streams to be converted into monitoring-oriented information that may support future phytoplankton observation and early-warning systems.





\bibliographystyle{elsarticle-num}

\bibliography{ref}

\end{document}